\documentclass[conference]{IEEEtran}
\IEEEoverridecommandlockouts
\usepackage{cite}
\usepackage{amsmath,amssymb,amsfonts}
\usepackage{algorithmic}
\usepackage{graphicx}
\usepackage{textcomp}
\usepackage{xcolor}
\usepackage{blindtext}
\usepackage[utf8]{inputenc}
\usepackage[colorinlistoftodos]{todonotes}
\usepackage{hyperref}
\hypersetup{colorlinks,allcolors=black}
\def\BibTeX{{\rm B\kern-.05em{\sc i\kern-.025em b}\kern-.08em
    T\kern-.1667em\lower.7ex\hbox{E}\kern-.125emX}}

\begin{document}

\newcommand{\will}{\textcolor{blue}}
\newcommand{\brian}{\textcolor{red}}

\title{Continual learning benefits from multiple sleep mechanisms: NREM, REM, and Synaptic Downscaling\\
\thanks{This project was funded by internal research and development funds from the Johns Hopkins University Applied Physics Laboratory.}

}

\author{
\IEEEauthorblockN{
Brian S. Robinson,
Clare W. Lau,
Alexander New,
Shane M. Nichols,\\
Erik C. Johnson,
Michael Wolmetz,
and William G. Coon
}
\IEEEauthorblockA{
The Johns Hopkins University Applied Physics Laboratory\\Research and Exploratory Development Department\\Intelligent Systems Center\\
Laurel, MD  USA
}
}


\maketitle
\thispagestyle{plain} \pagestyle{plain}

\begin{abstract}
Learning new tasks and skills in succession without overwriting or interfering with prior learning (i.e., “catastrophic forgetting”) is a computational challenge for both artificial and biological neural networks, yet artificial systems struggle to achieve even rudimentary parity with the performance and functionality apparent in biology. One of the processes found in biology that can be adapted for use in artificial systems is sleep, in which the brain deploys numerous neural operations relevant to continual learning and ripe for artificial adaptation.
Here, we investigate how modeling three distinct components of mammalian sleep together affects continual learning in artificial neural networks: (1) a veridical memory replay process observed during non-rapid eye movement (NREM) sleep; (2) a generative memory replay process linked to REM sleep; and (3) a synaptic downscaling process which has been proposed to tune signal-to-noise ratios and support neural upkeep.  
To create this tripartite artificial sleep, we modeled NREM veridical replay by training the network using intermediate representations of samples from the current task. We modeled REM by utilizing a generator network to create intermediate representations of samples from previous tasks for training. Synaptic downscaling, a novel contribution, is modeled utilizing a size-dependent downscaling of network weights.
We find benefits from the inclusion of all three sleep components when evaluating performance on a continual learning CIFAR-100 image classification benchmark. Maximum accuracy improved during training and catastrophic forgetting was reduced during later tasks.
While some catastrophic forgetting persisted over the course of network training, higher levels of synaptic downscaling lead to better retention of early tasks and further facilitated the recovery of early task accuracy during subsequent training.
One key takeaway is that there is a trade-off at hand when considering the level of synaptic downscaling to use -  more aggressive downscaling better protects early tasks, but less downscaling enhances the ability to learn new tasks. Intermediate levels can strike a balance with the highest overall accuracies during training. 
Overall, our results both provide insight into how to adapt sleep components to enhance artificial continual learning systems and highlight areas for future neuroscientific sleep research to further such systems.

\end{abstract}


\section{Introduction}

Learning new tasks and skills in succession without overwriting or interfering with prior learning (i.e., “catastrophic forgetting”) is a computational challenge for both artificial and biological neural networks. And yet, while the latter overcome this with ease (ex., a child does not forget what s/he learned in class yesterday by learning new things today), current artificial networks struggle to achieve parity. Several approaches for this kind of "Continual Learning" problem \cite{parisi2019continual} have been developed in the A.I. space, including: dynamic architectures that can grow network capacity, weight regularization-based approaches that mitigate catastrophic forgetting by constraining the update of previous weights \cite{kirkpatrick2017overcoming,zenke2017continual}, and interleaved replay of training examples from previous tasks \cite{hayes2021replay}.  In contrast to rote replay (e.g., using a memory buffer to store one-to-one copies of experience),
 generative replay \cite{hayes2021replay} does not store exact copies of specific examples from previous tasks, but instead trains a network to retain higher-level/compressed representations, from which it can create \textit{de novo} synthetic training samples.  

A general challenge with all strategies for continual learning
lies in their ability to scale. Dynamic architectures and replay approaches must ensure that the network's (or replay buffer) size remains manageable.
Generative replay approaches specifically can produce ever-growing generator networks that are susceptible to catastrophic forgetting.
Weight regularization does not extend well to more challenging tasks like class-incremental learning (which requires learning from one subset of classification targets to persist in the face of constant re-training on incrementally presented new subsets) \cite{hsu2018re}.  Regularization methods may not scale to large numbers of tasks, as weight regularization parameters continue to grow.  

Although continual learning clearly poses daunting challenges for artificial systems, given the abundance of \textit{biological} solutions that can be adapted into artificial ones, there is ample opportunity to improve artificial systems by leveraging these as blueprints. 

Mammalian brains in particular have evolved a wide array of specialized processes to combat catastrophic forgetting, and several of the operations that occur most saliently during sleep are attractive candidates for artificial implementation. These include: 1) a veridical memory replay process linked to non-rapid eye movement (NREM) sleep; 2) a generative memory replay process linked to REM sleep; and 3) a synaptic downscaling process which has been proposed to modulate signal-to-noise ratios and support neural upkeep \cite{tononi2014sleep}.  Much is understood about these processes from a neuroscience perspective, and it is worth briefly reviewing the neuroscience behind these processes to shed light on their appealing properties and provide intuition about their operating principles. 

One of the most widely studied and ubiquitous learning processes visible in the brain during sleep is memory replay. Memory replay has been observed during sleep in animals \cite{wilson1994reactivation,skaggs1996replay} and in humans \cite{schreiner2021endogenous}, and is perhaps most convincingly demonstrated in the hippocampus of rodents engaged in spatial learning (hippocampus-dependent) tasks.  These studies leverage the useful properties of hippocampal place cells, neurons that fire vociferously when the rodent is located in that cell's receptive field \cite{o1971hippocampus}.  Together, a sequence of place cell firings encodes a specific trajectory through space, and trajectories observed during waking behavior can be detected in subsequent bouts of sleep via correlation analyses \cite{skaggs1996replay}.  In this way, the neural representation of specific information known to the experimenter and encoded in the animal's neural circuitry can be precisely quantified and studied during sleep.

Memory replay during NREM sleep connects novel learning supported by the hippocampus to distant afferents in the cortex \cite{latchoumane2017thalamic}, and has been proposed to support the consolidation of long-term memory into distributed cortical stores \cite{latchoumane2017thalamic,siapas1998coordinated}.  In NREM sleep, replay is \textit{veridical}, meaning, place cell trajectories activated during learning are replayed in exactly the same sequences as those observed during wakefulness \cite{lee2002memory}.  Veridical replay facilitates network plasticity in mammalian brains. Likewise, in artificial ones, it could be exploited to update network weights several times over rather than only during initial exposure to the represented experience/example.  This type of process can also be implemented by generating samples at intermediate neural network layers instead (i.e., restricting operations to only higher-level, abstract representation layers, rather than fully recreating synthetic examples for submission to early input layers; this is effective while furthermore being more efficient).\cite{van2020brain}.  


\begin{figure}
    \centering
    \includegraphics[width=.97\columnwidth]{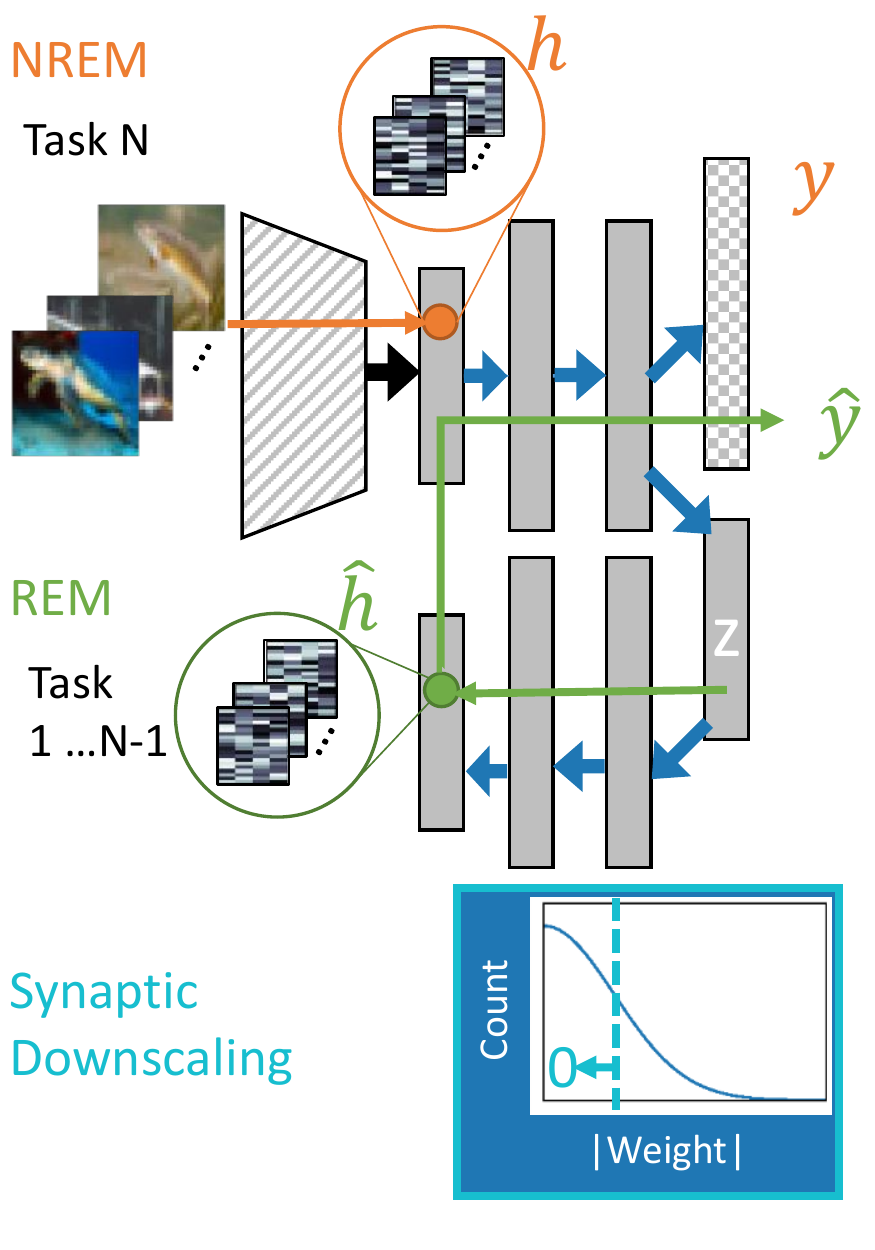}
    \caption{Tripartite artificial sleep modeling NREM, REM, and synaptic downscaling. A network model contains a pre-trained set of convolutional layers (hatched), a symmetrical VAE (solid), and a classifier output layer (checkered) projecting from the final VAE encoding layer. NREM veridical replay is modeled by sampling from intermediate processed image features ($h$) and class targets ($y$) for the current task, $N$. REM generative replay is modeled from previous tasks by sampling the latent stochastic variable layer, $z$, to generate intermediate feature, $\hat{h}$, which are provided as input to the VAE encoder to generate class target estimates $\hat{y}$. All trainable weights (blue) are updated for the VAE and classifier output utilizing the replayed samples (in $h$ and $\hat{h}$) and class targets ($y$ and $\hat{y}$). For synaptic downscaling, size-dependent scaling is modeled by setting weights below a specified percentile to zero for each layer prior to training for each task. }
    \label{fig:Tripartite}
\end{figure}


In stark contrast to the hippocampo-cortical outflow seen during NREM sleep, during REM sleep, the flow of information is reversed and hippocampal output to the cortex is suppressed \cite{chrobak1994selective,hasselmo1996suppression}.  This frees the cortex to process and reorganize knowledge without interference not only from the environment but also from the hippocampus \cite{stickgold2000visual}. In mammals, this is likely achieved mechanistically by REM's distinctive neuromodulatory tone (low acetylchline and norepinephrine) which favors the spread of neuronal activity beyond that observed during wakefulness \cite{hasselmo1999neuromodulation}. This increased connectivity and opportunity for novel connections provides a possible physiological basis for the finding that sleep generates insight in humans \cite{wagner2004sleep}, which in turn alludes to another benefit that could be tapped by artificial analogues. 
Viewed from the perspective of continual learning, REM sleep is a state ideally suited to \textit{generate} novel possible experiences by manipulating and reorganizing elements experienced over the lifetime.  This generative replay likely supports further memory consolidation in mammals, and presents a mechanism for artificial neural networks to revisit representations of experience learned in the distant past and generate novel feature combinations that share the statistical properties of previously experienced examples.



A fundamental sleep process that remains to be explored in artificial neural networks is synaptic downscaling. Synaptic downscaling has been observed in animals to be size-dependent (i.e., scaled by the size of interfacing synaptic boutons \cite{de2017ultrastructural}), and has been hypothesized to homeostatically regulate the metabolic costs incurred by synaptic connections formed during wakefulness, and recycle unneeded synapses for future use\cite{tononi2003sleep,tononi2014sleep}. Superficially a purely metabolic function, this process has been proposed to have the critically important effect of fine-tuning signal-to-noise (SNR) ratios in neural networks. Regulating weight updates in a similar way may be a useful addition to artificial neural networks tasked with continual learning scenarios.

There have been other approaches for continual learning that utilize aspects of sleep, such as implementing oscillatory phase coding of unsupervised spike-timing-dependent plasticity during training \cite{krishnan2019biologically}, modeling hippocampal consolidation into a medial prefrontal cortex generator model \cite{Kemker2018}, and creating detailed thalamo-cortical models of replay and slow-wave sleep for sequence learning \cite{gonzalez2020can}.  However, the integration of a synaptic downscaling process into an artificial neural network that also implements two types of memory replay (veridical and generative) has not yet been investigated. 

To address this gap, in this work we train an artificial neural network on a continual learning task that includes models of all three of these sleep processes: 1) NREM veridical memory replay, 2) REM generative replay, and 3) synaptic downscaling. To address the first, NREM veridical replay is modeled by interleaving processed sensory input examples from the current task during training. For the second, REM generative replay, we incorporate a model of replay\cite{van2020brain} that generates statistically matched composites of processed sensory input features experienced in previous tasks to use during continued training.  The model learns latent representations of object classes, from which it can further learn by generating and evaluating the composed results. Third and finally, to model the size-dependent downscaling of synaptic boutons that has been observed during biological sleep\cite{de2017ultrastructural}, we incorporate magnitude-based pruning as a first order approximation of the process. Magnitude-based pruning not only provides a simplified downscaling approach with few hyperparameters, but has also been shown to be generally effective for model compression (even compared to more complex sparsity-inducing methods) \cite{gale2019state}. 
Combining all three of these facets of sleep is, to the best of our knowledge, a novel approach to continual learning, and alludes to more gains that may be realized by adopting this biofidelic approach.

In sum, in this study we investigate the joint performance of three sleep-inspired neural processes implemented in a neural network that trains on a challenging CIFAR-100 class-incremental continual learning benchmark task. We then evaluate the network's ability to perform image class prediction from any of the sets of previously trained tasks (i.e., the sum total of its ``lifetime'' experience). In light of the observed performance gains, our results indicate that not only is synaptic downscaling a useful approach, but that in general adopting this frame of reference, i.e., turning to mammalian sleep and modeling the multiple neural processes that occur there, is a useful perspective to assume when seeking new ways to improve the performance of future artificial neural networks and intelligent systems. 

\section{Methods}


An initial model of tripartite artificial sleep for network training is implemented by extending and integrating approaches that have been used separately for continual learning and neural network model compression.
While there are many types of continual learning benchmark tasks, \textit{class incremental learning}, wherein a network needs to classify examples from \textit{any} previously learned task (as opposed to performing classification for a single, previously learned task) is more challenging than other scenarios (e.g., incremental learning\cite{van2020brain}) and more akin to human learning, which occurs across a lifetime (i.e., the sum total of experience)).  Hence we selected this approach to use here.

A network implementing tripartite artificial sleep (Fig. \ref{fig:Tripartite}), needs to be able to generate past training examples (replay) as well as be able to perform classification on the current task.
In this work, a model of the NREM veridical replay process is implemented by utilizing feature representations in intermediate network layers of the current task's training data for weight optimization.
A model of REM generative replay is implemented as generative ``hidden'' replay, where a generator / auto-encoder generates training input samples and the classifier output is used for REM training labels.
The classifier training is performed using the NREM / REM input samples and class labels, while the generator training is performed using the NREM / REM input samples in tandem with classifier training.
Synaptic downscaling during the sleep / wake cycle has been observed in animals to be size-dependent, and this is modeled here in a first order approximation of size-dependent scaling (setting a cutoff threshold for weight zeroing). With the implemented synaptic downscaling, small weights are down-scaled completely (zeroed out) and larger weights are not down-scaled at all. This is implemented once on each new task (set of ten classes) at the beginning of training (on that task).



The two-process (veridical/generative replay) base network architecture that we modified for tripartite artificial sleep has been previously used to investigate continual learning \cite{van2020brain}, and consists of (1) a set of five pre-trained convolution layers that take raw images as input and output a vector of image features, $\mathbf{h}$,  (2) a symmetric variational autoencoder (VAE), which consists of (a) an encoder network that maps $\mathbf{h}$ to a vector of stochastic latent variables, $\mathbf{z}$, and (b) a decoder network which generates an estimated reconstructed image feature vector $\mathbf{\hat{h}}$, and (3) a softmax classification output layer, which receives input from the last layer of the VAE encoder network (Fig. \ref{fig:Tripartite}).
The five convolution layers have 16, 32, 64, and 254 channels respectively with 3$\times$3 kernels and a padding of 1. All layers have a stride of 2, except for the first layer which has no downsampling. The input into the convolution layers is a 32$\times$32 RGB image, and the output, $\mathbf{h}$,  is vector of 1,024 flattened image features.
The encoder and decoder VAE networks each consist of two fully-connected layers of 2,000 ReLU units. The stochastic latent variable layer, $\mathbf{z}$, has 100 Gaussian units.
The softmax output layer has a unit for each class label to be predicted.
For the split CIFAR-100 class incremental continual learning task and replay-based optimization \cite{van2020brain}, the CIFAR-100 dataset \cite{krizhevsky2009learning} is split into 10 tasks with 10 classes each.
For training, the loss function, $\mathcal{L}$, that is optimized during a task, 
$$\mathcal{L}=\mathcal{L}^C+\mathcal{L}^G,$$
is a combination of classification loss, $\mathcal{L}^C$, and generator loss, $\mathcal{L}^G$.
Prior to training for task $N$, the stochastic latent variable, $\mathbf{z}$, is sampled to generate image feature samples $\mathbf{\hat{h}}_j$, $j \in \{1...N-1\}$ from previous tasks. Corresponding classifier softmax output samples $\mathbf{\hat{y}}_j$ are generated by passing $\mathbf{\hat{h}}_j$ as input to the encoder network. 
The classification loss is composed of $$ \mathcal{L}^C=\frac{1}{N}\mathcal{L}^C_{current} + (1-\frac{1}{N})\mathcal{L}^C_{replay},$$ where $\mathcal{L}^C_{current}$ is the cross entropy loss calculated for the current task and ${L}^C_{replay}$ is the distillation loss calculated for samples from the previous tasks.
The generator loss is composed of 
$$ \mathcal{L}^G=\frac{1}{N}\mathcal{L}^G_{current} + (1-\frac{1}{N})\mathcal{L}^G_{replay}, $$
where ${L}^G_{current}$ is calculated based on image features, $\mathbf{h}$, from the current task images and ${L}^G_{replay}$ is calculated based on generated image features samples $\mathbf{\hat{h}}_j$. Optimization is performed for 10,000 iterations per task with the ADAM-optimizer ($\beta_1=0.9, \beta_2=0.999$).
The convolutional layers were pre-trained on a classification task with non-overlapping images (CIFAR-10).

A simplified model of synaptic downscaling introduced here is incorporated into network training by setting a fraction, $p$, of the smallest weights in each trainable layer to zero for each task prior to weight optimization. To gain insight into the functional importance of the modeled sleep processes, network training variants are evaluated with varying levels of $p \in \{0, 0.25, 0.5, 0.75, 0.9\}$, as well as with and without the modeled REM generative replay.  

\section{Results}

\begin{figure*}
    \centering
    \includegraphics[width=.8\textwidth]{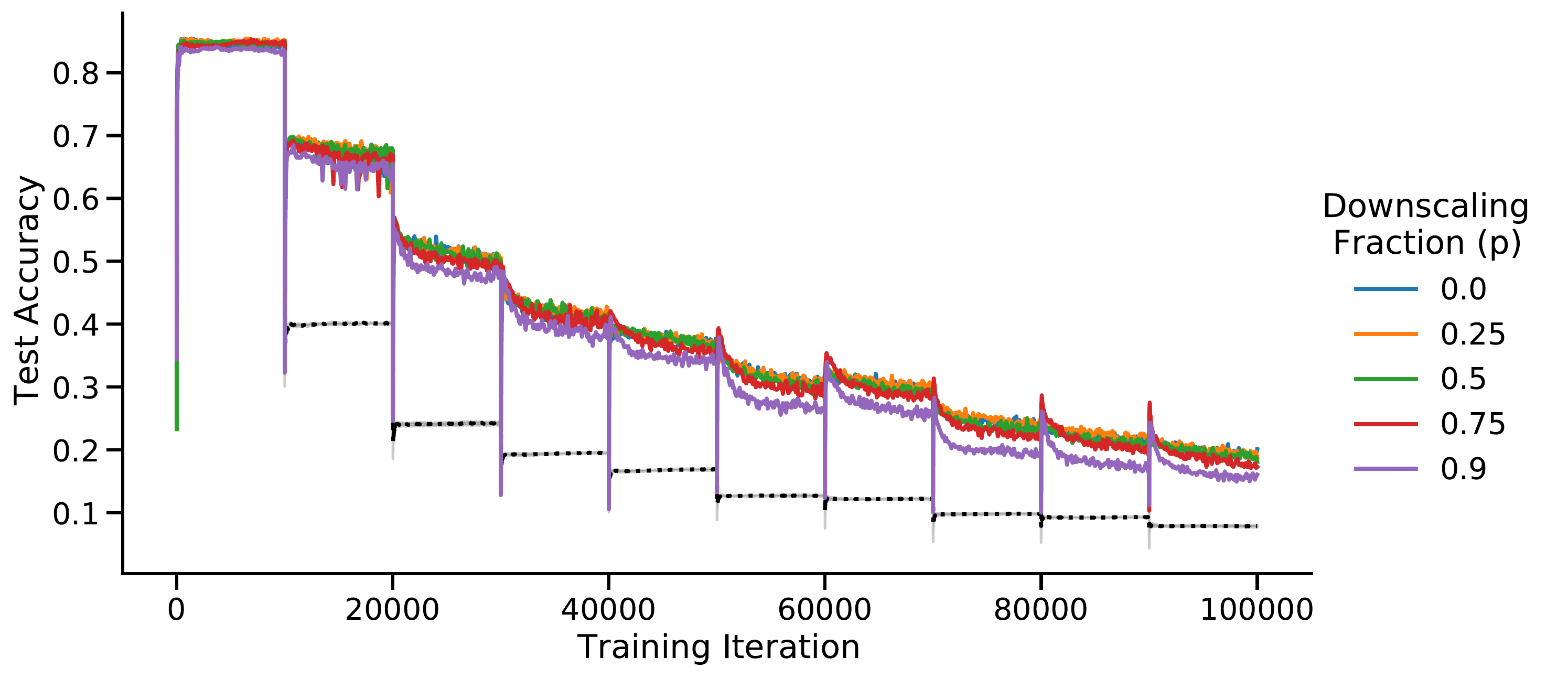}
    \caption{Continual learning average accuracy during training with varying REM generative replay and different levels of synpatic downscaling. The dashed black line demonstrates the collapse of average testing accuracy across all evaluated synaptic downscaling levels without REM generative replay (with bootstrapped 95\% confidence intervals). The colored lines demonstrate the average test accuracy with REM generative replay and different levels of synaptic downscaling.}
    \label{figure.fig2}
\end{figure*}

\begin{figure}
    \centering
    \includegraphics[width=\columnwidth]{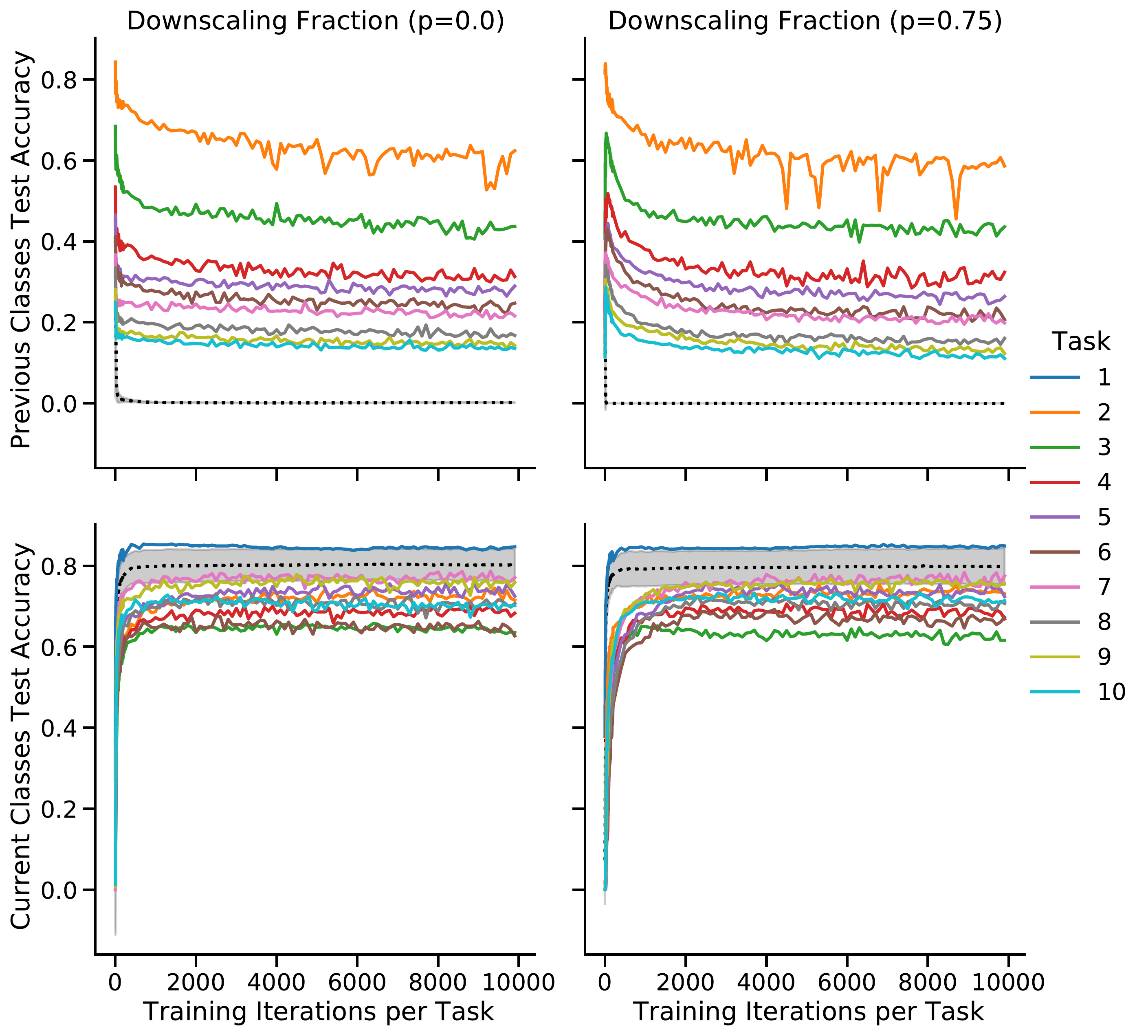}
    \caption{Comparison of task performance in the current versus previously trained tasks. Top and bottom panels represent replayed and current task accuracy respectively. Left and right panels depict the absence and presence of downscaling. The dashed black line indicates average testing accuracy across all evaluated synaptic downscaling levels without REM generative replay (with bootstrapped 95\% confidence intervals). The colored lines demonstrate the average test accuracy with REM generative replay and different levels of synaptic downscaling.}
    \label{figure.fig3}
    
\end{figure}

For the class-incremental continual learning benchmark, as the task number increases, task difficulty increases both because there are more valid classifier outputs and because there have been more iterations of training updates since early task images have been provided to the network.
An overview of testing accuracy across different components of tripartite artificial sleep (Fig. \ref{figure.fig2}), highlights the degradation in model accuracy when no generative replay is utilized. The average testing accuracy, $\mu_N(i)$, is quantified for each training iteration, $i$, during the current task, $N$ as $$ \mu_N(i) = \frac{1}{N}  \sum_{C=1}^{N}  a_N^C(i),$$ where $a_N^C(i)$ is the test accuracy for the current task, $N$, on the set of classes introduced during task $C$.  
Without generative replay, the accuracy on the current task is high, but there is a complete collapse on accuracy from previous tasks (Fig. \ref{figure.fig3}). The average testing accuracy on the previous classes is measured as $\mu^{prev}_N(i) =  \frac{1}{N-1}  \sum_{C=1}^{N-1}  a_N^C(i)$ and correspondingly the average testing accuracy on the current task is measured as $\mu^{current}_N(i) = a_N^N(i)$.
With generative replay, model accuracy is increased relative to applying only veridical replay, however, during the course of task training, the overall testing accuracy still decreases (Fig. \ref{figure.fig2}). The decrease in overall accuracy during training occurs because even though the accuracy on classes introduced during the current task increases, the accuracy on classes introduced during previous tasks invariably decreases.
With synpatic downscaling, one of the most striking observed trends is that for higher downscaling levels on later tasks, there is a prolonged period of higher replayed accuracy (Fig. \ref{figure.fig3}).
Overall, analysis of the full trajectory of testing accuracy during training demonstrates that generative replay is necessary to mitigate catastrophic forgetting, and that the level of downscaling affects the progression of maintained accuracy on previous tasks such that later tasks are maintained for longer.

\begin{figure}
    \centering
    \includegraphics[width=\columnwidth]{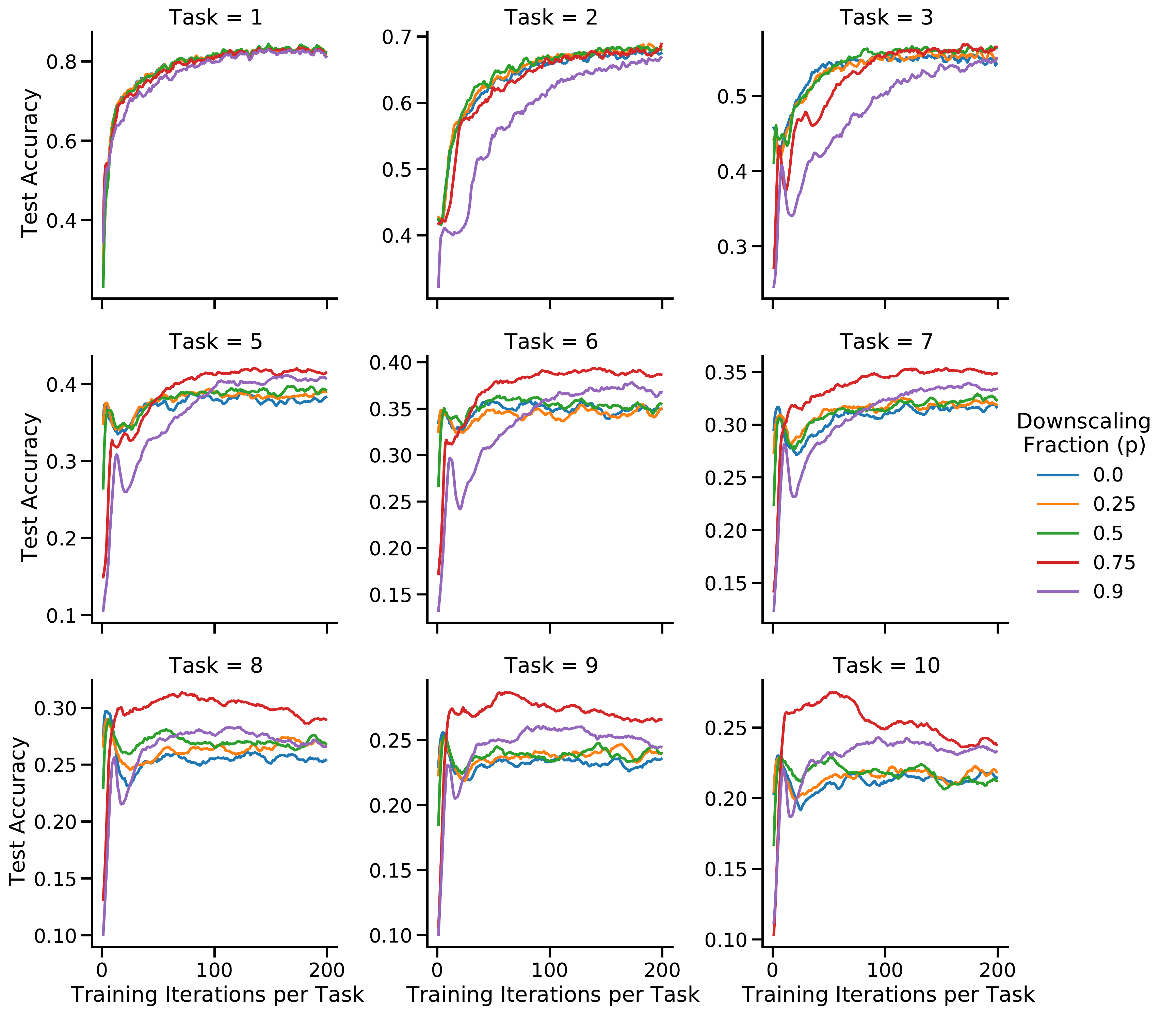}
    \caption{Dynamics of test accuracy during early training iterations when varying the fraction of downscaling. With $p=0.75$ in later tasks, higher test accuracies are observed as well as a shifting of the peak values to later training iterations.}
    \label{figure.fig4}
\end{figure}

\begin{figure}
    \centering
    \includegraphics[width=.8\columnwidth]{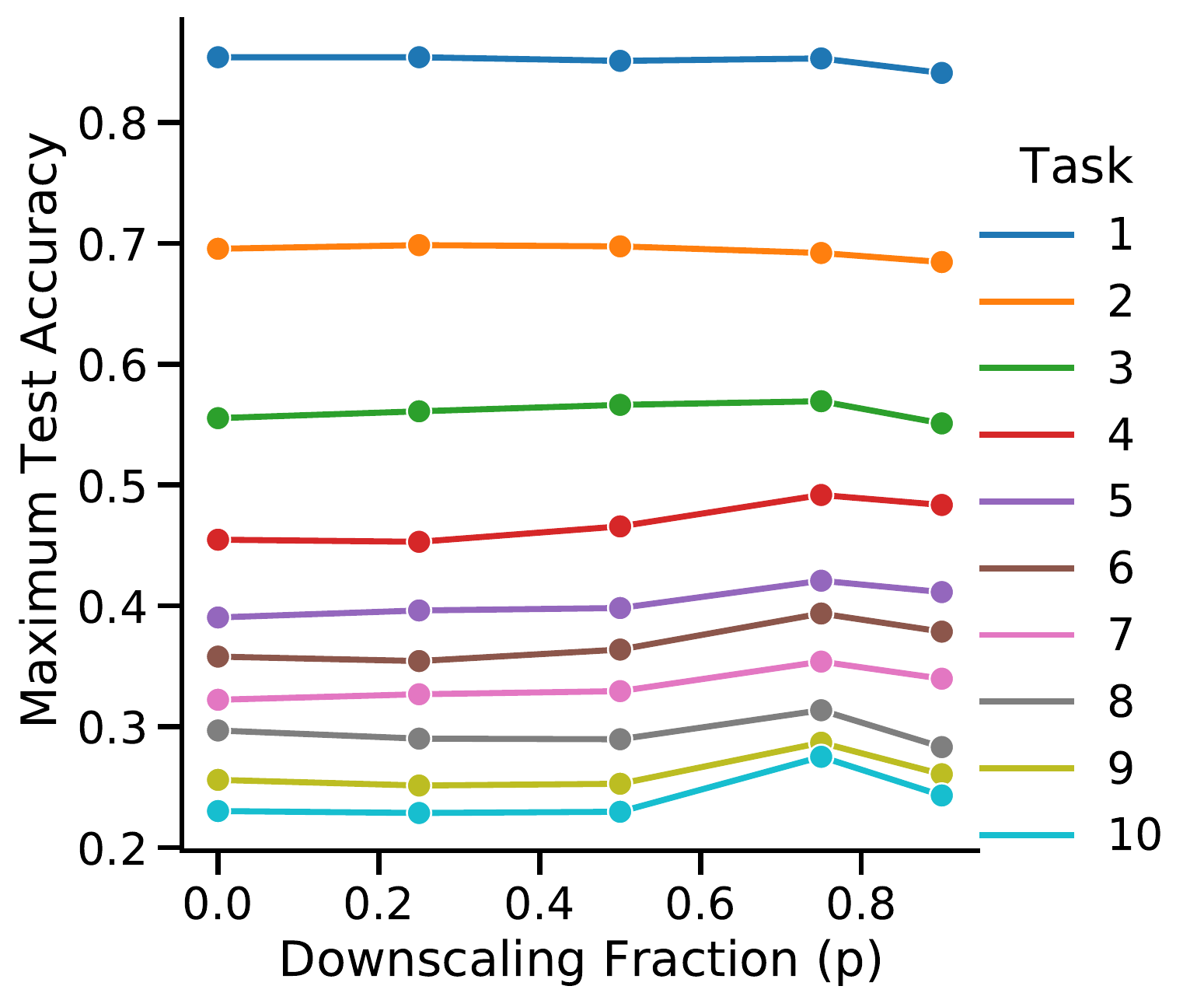}
    \caption{Maximum overall testing accuracy during training is influenced by synaptic downscaling. Maximum test accuracy is measured as the maximum of accuracy, $\mu_N(i)$, across all training iterations for each task (always measured on all introduced classes). For later tasks, maximum test accuracies are observed for a downscaling fraction of $p=0.75$. }
    \label{figure.fig5}
\end{figure}

In all of the training variants with generative replay after the first task, there is a rapid increase in overall accuracy (Fig. \ref{figure.fig4}) followed by a steady decline.  
These dynamics of testing accuracy over the course of training iterations changes per downscaling level (Fig. \ref{figure.fig4}). Without any downscaling or with lower levels of downscaling ($p<=0.5$), the trajectory looks similar.
For later tasks, with increasing $p$, the accuracy continues to climb for longer, with a later peak in overall accuracy.
With $p=0.75$, there is a clear increase in overall accuracy at later tasks (Fig. \ref{figure.fig4}-\ref{figure.fig5}). The maximum test accuracy during training of each task (maximum of $\mu_N(i)$) can be used as a metric to summarize overall performance. For early tasks, the maximum accuracy is relatively preserved, however, for later tasks, there is a trend for maximum accuracy to increase up to a downscaling level of $p=0.75$ and decrease as downscaling levels increase further.
The maximum accuracy levels per evaluated downscaling level can be understood in the context of the trade-off between current and replayed task accuracy. During training without downscaling, there is a rapid decay of replayed accuracy, which occurs before a sufficient increase in current task accuracy, leading to lower maximum accuracy levels. During training with downscaling $p=0.75$, there is a prolonged maintenance of replayed accuracy that overlaps with an increase in current task accuracy, leading to a higher maximum accuracy level. 

\begin{figure}
    \centering
    \includegraphics[width=\columnwidth]{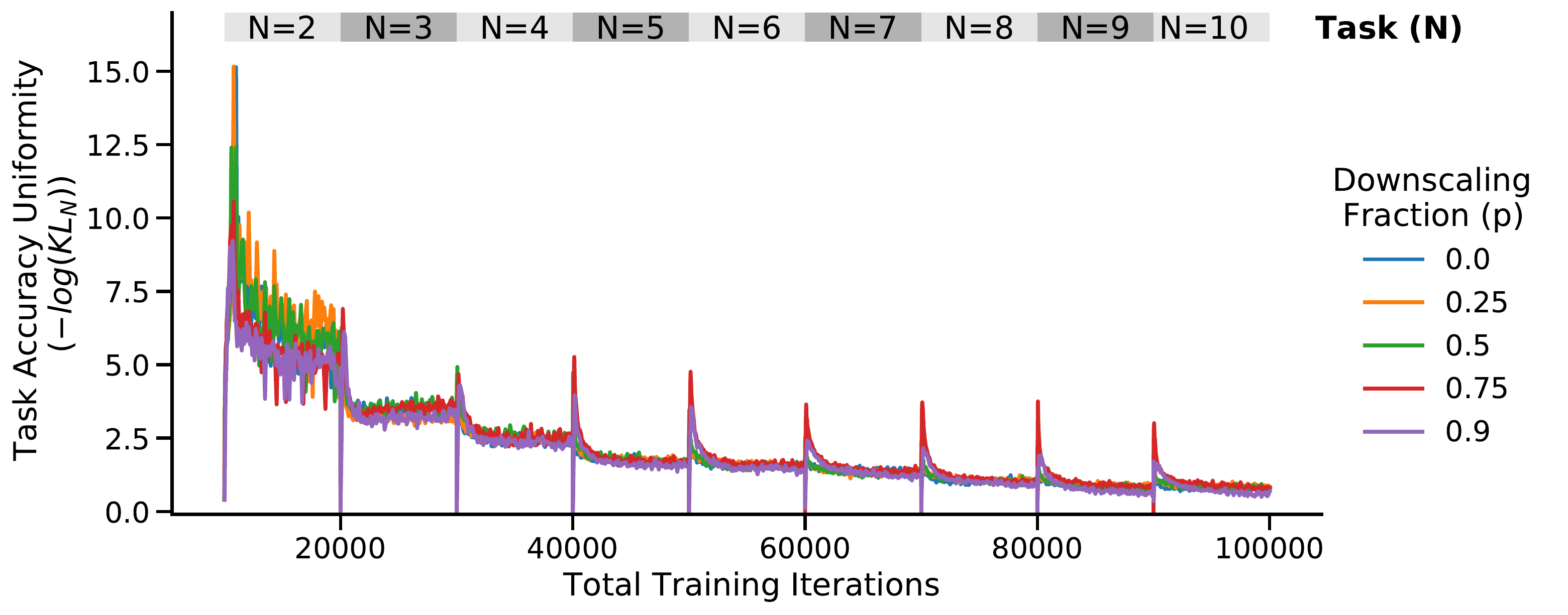}
    \caption{Uniformity of accuracy across task classes. Higher values indicate that the distribution of accuracy across task classes is more balanced and closer to a uniform distribution as measured with KL-divergence.  }
    \label{figure.fig6}
\end{figure}
\begin{figure}
    \centering
    \includegraphics[width=\columnwidth]{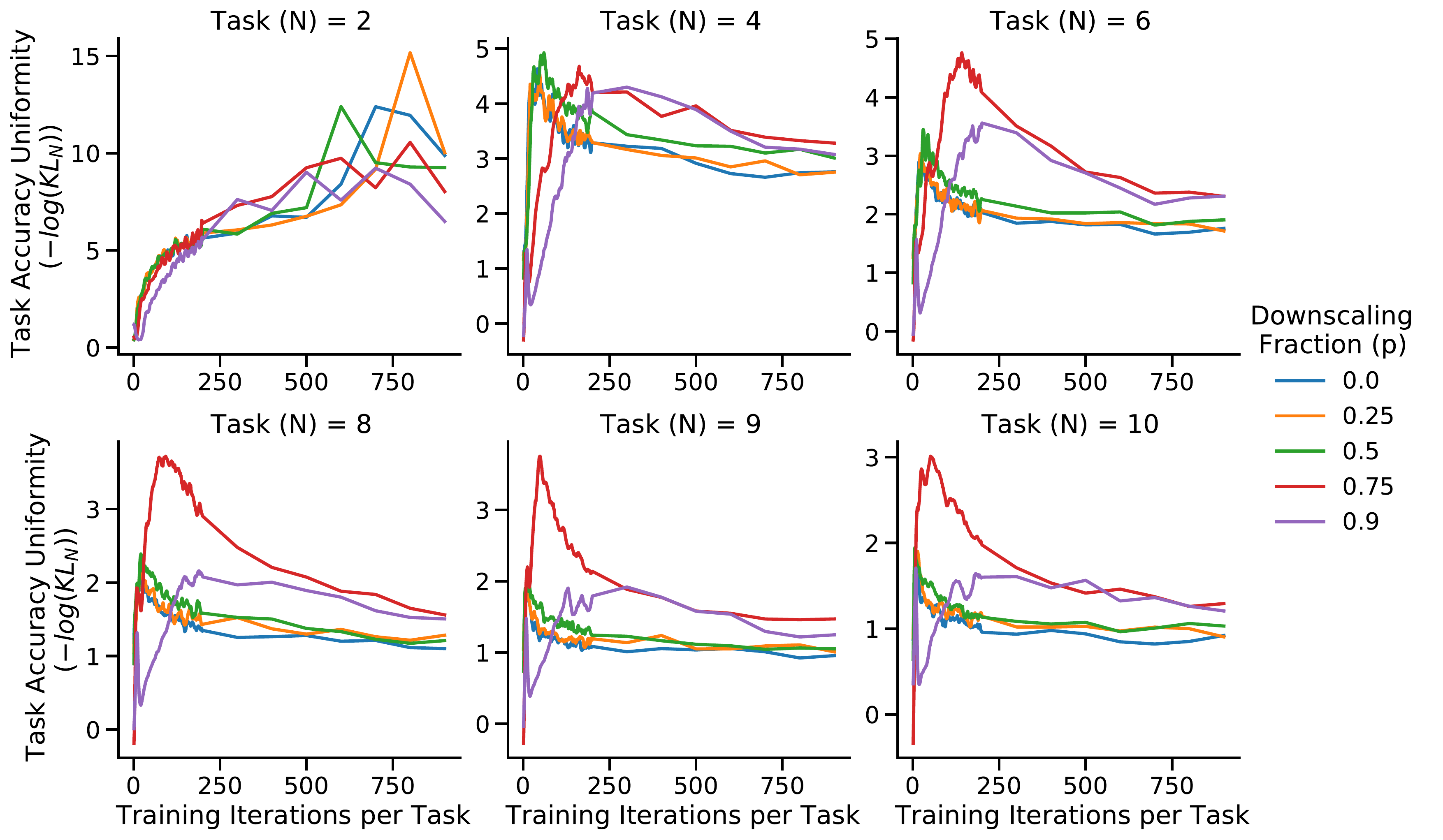}
    \caption{Uniformity of accuracy across task classes during early training iterations. For later tasks with downscaling at $p=0.75$, task accuracies are more balanced across classes (the observed distribution is closer to a uniform distribution as measured with KL-divergence).}
    \label{figure.fig7}
\end{figure}
\begin{figure}
    \centering
    \includegraphics[width=\columnwidth]{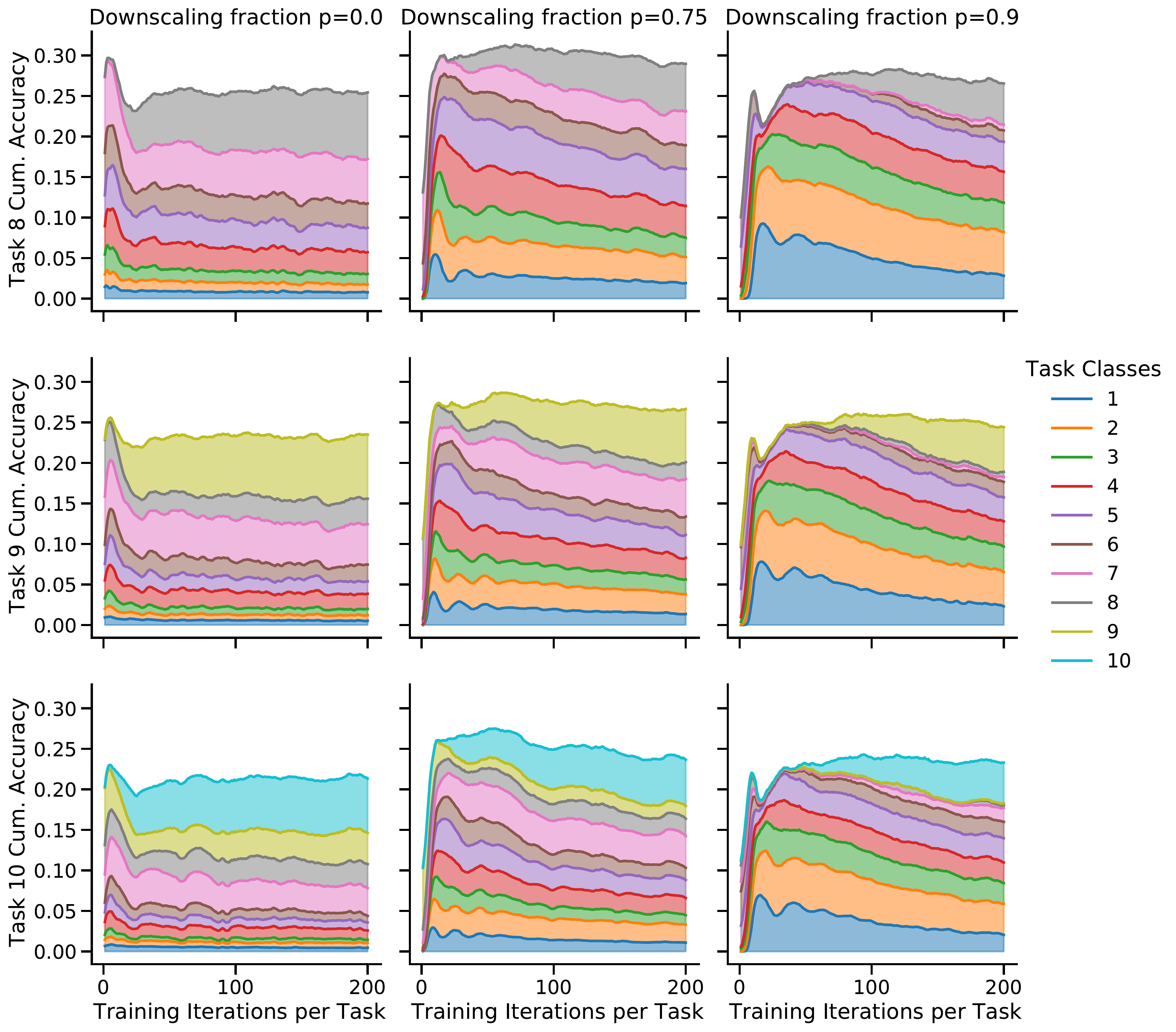}
    \caption{Distribution of accuracy across task classes during early training iterations. Higher levels of synaptic downscaling better protect earlier learned classes against forgetting. The height of all colored regions represents the overall average accuracy across all tasks ($\mu_N$), while the height of each individual colored region represents the accuracy ($a_N^C$) for classes introduced during task $C$ normalized by the number of colored regions, $N$.}
    \label{figure.fig8}
\end{figure}


\begin{figure*}
    \centering
    \includegraphics[width=.8\textwidth]{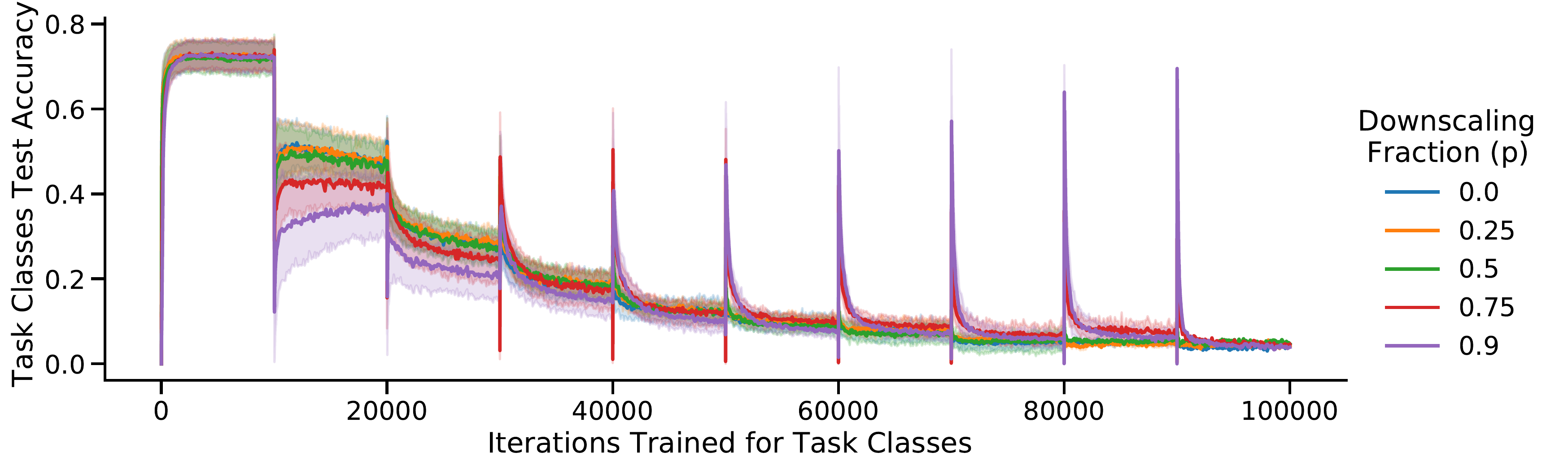}
    \caption{Forgetting characterization. As the amount of total training iterations after a class is first introduced increases, the task class accuracy diminishes (forgetting occurs). However, there is a period of recovered accuracy which is higher with more synaptic downscaling. Solid lines are the average across all sets of task classes with bootstrapped 95\% confidence intervals. Note that there have been less total iterations trained for task classes that are introduced in later tasks, thus less accuracies are included for task average accuracies as iterations trained for task classes increases.}
    \label{figure.fig9}
\end{figure*}

\begin{figure}
    \centering
    \includegraphics[width=\columnwidth]{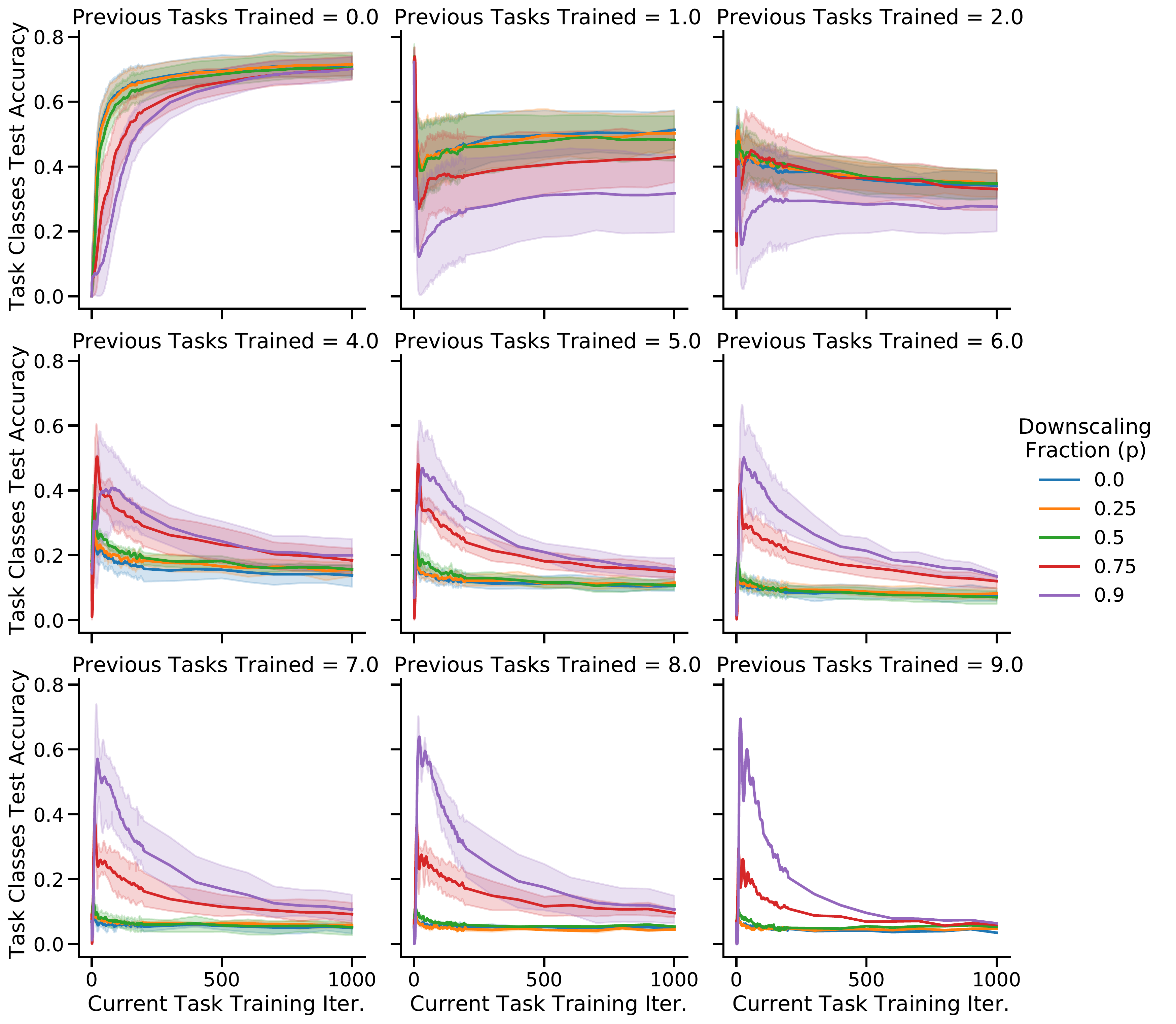}
    \caption{Forgetting characterization during early training iterations. After more successive tasks are trained, class accuracy recovery becomes pronounced with more synaptic downscaling.}
    \label{figure.fig10}
\end{figure}

We can evaluate the properties of a continual learning system by further understanding how task performance is affected by the makeup of performance across previous tasks.
A system that learns in a way that is balanced between all tasks would have uniform task accuracy between tasks, which we can measure during each training iteration as the KL divergence of the observed accuracy across tasks and a uniform distribution, 

$$ KL_N(i) =   \sum_{C=1}^{N} \frac{a_N^C(i)}{N \mu_N(i)} log( \frac{a_N^C(i)}{\mu_N(i)} ) $$ 

Smaller $KL_N(i)$ values signify a balance of accuracies.
Generally, over the course of training epochs for a task, there is a rise in $KL_N(i)$ as the current task accuracy dominates over the previous tasks (Fig. \ref{figure.fig6}). In later tasks, however, there is a minimum $KL_N(i)$ value (signifying balance between tasks) with downscaling at $p=0.75$ (Fig. \ref{figure.fig7}). 
Underscoring the importance of balance between tasks, the training iterations with minimum $KL_N(i)$ value overlaps with the training epochs for the highest overall accuracy.
Note that there is another trend of instantaneous rise and drop of $KL_N(i)$ right after downscaling which can be attributed to the discontinuity after downscaling.
The balance of task accuracies can be further understood by investigating the relationship between recency and task balance (Fig. \ref{figure.fig8}). Without downscaling, task accuracy is driven predominantly by the most recent tasks with diminishing contributions from earlier tasks. Contrarily, with extreme downscaling ($p=0.9$), overall accuracy is driven by earlier tasks (at the expense of more recent tasks). At the intermediate value of $p=0.75$, the task accuracy between previous tasks is balanced.
Overall, we find that the level of downscaling affects the balance of accuracy between previous tasks, where no downscaling overrepresents later tasks, extreme downscaling overrepresents earlier tasks, and intermediate levels of downscaling maintains a more balanced level of accuracy between tasks.

To enable a system to have higher overall accuracy, forgetting of previous tasks needs to be limited.
We can quantify the forgetting of a set of classes introduced in task $C$ during training as $$ f^m_C(i) = a_{m+C}^C(i),$$ where $m$ is the number of additional tasks that have been introduced after $C$. (Fig. \ref{figure.fig9}). When $m=0$, $f^m_C$ signifies the accuracy when the set of classes, $C$ is first being introduced, where an overall increase in task accuracy is observed. During training on subsequent tasks, a decline in accuracy is observed.
A striking trend, however, is that with downscaling, there is a pronounced recovery in previous task accuracy, with accuracies on initially trained tasks capable of reaching over 60\% accuracy (Fig. \ref{figure.fig10}).
The level of downscaling affects this recovery, with larger amounts of downscaling resulting in larger recovery of accuracies.
The relationship between downscaling and forgetting bolsters what has been seen in the task balance and cumulative accuracy, where downscaling diminishes forgetting of earlier tasks. 
Even though the most extreme levels of downscaling has the most protection of performance on earlier tasks, there is still a trade off, where the protection of earlier tasks competes with the ability to learn newer tasks. Thus high, but not extreme downscaling may be preferable in practice.

\begin{figure}
    \centering
    \includegraphics[width=\columnwidth]{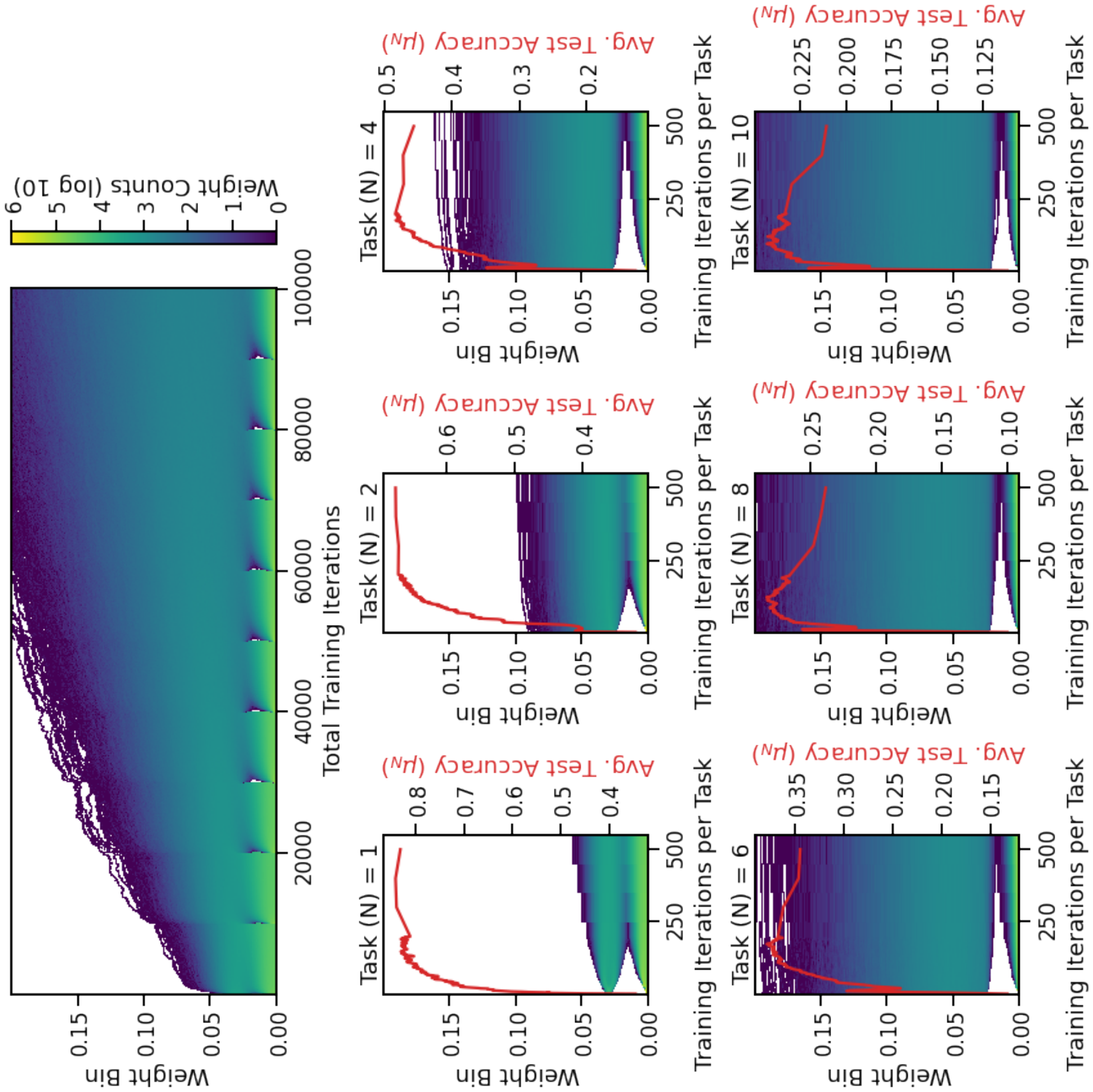}
    \caption{Distribution of positive network weights evolves over the course of training across tasks with synaptic downscaling. Top) Network weights in the first encoding layer with downscaling fraction $p=0.75$ broadens over the course of training. Bottom) During early training iterations of above, the bi-modal structure of network weights after downscaling diminishes. In later tasks, the test accuracy decreases as the network weights approach a less bi-modal distribution.}
    \label{figure.weight_dists}
\end{figure}

\begin{figure}
    \centering
    \includegraphics[width=\columnwidth]{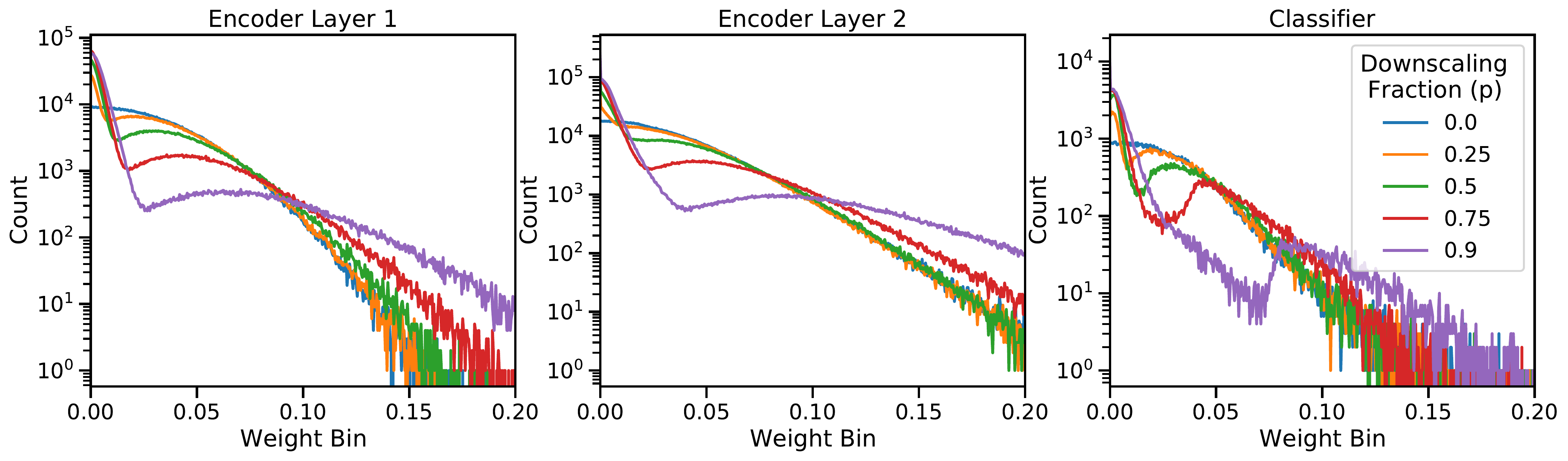}
    \caption{Final positive weight distribution in network layers with different levels of synaptic downscaling. Distributions with more downscaling are broader and more bi-modal.}
    \label{figure.final_Weights}
\end{figure}

To further understand the mechanistic relationship between downscaling and system performance, we investigate how the network's distribution of weights changes over the course of training (Fig. \ref{figure.weight_dists}).
Generally, the overall weight distribution widens over the course of training on all tasks.
After downscaling, there is a bi-modal distribution of weights between the preserved and the downscaled weights, which becomes more uni-modal over the course of training and weight update.
For later tasks, the peak in overall accuracy corresponds to the training periods where the weight distribution is more bi-modal.
The amount of downscaling utilized also affects the overall distribution of network weights. With more downscaling, there is a broadening of the overall weight distribution with a more pronounced bi-modal structure.
The comparisons of trends in distributions with downscaling in artificial networks could be compared to the distributions observed in biology to better constrain future continual learning systems.

\section{Discussion}

In this work, we investigated the impact of a tripartite artificial sleep model in the training of an artificial neural network performing a continual learning task. This entailed three processes created to capture their respective beneficial properties observed in mammalian sleep: 1) NREM veridical replay, 2) REM generative replay, and 3) synaptic downscaling. 
We found that the addition of synaptic downscaling complements replay by enhancing continual learning and mitigating catastrophic forgetting.
Specifically, during training, the addition of synaptic downscaling was found to enhance performance on earlier tasks, increase the balance of accuracy between previous tasks and recent tasks, and achieve the highest overall accuracy. 
Furthermore, while early task accuracy diminishes over the course of network training, the inclusion of synaptic downscaling increases the recovery of early task accuracy during subsequent training.

Our findings, specifically that the addition of this third component (synaptic downscaling) improves continual learning, lend themselves to several interpretations.  One potential mechanism behind the overall increased performance is that downscaling implements model compression, similar to magnitude-based pruning \cite{zhu2017prune}, such that downscaled weights can be preferentially utilized on subsequent tasks.
Unlike pruning for model compression applied to a single task, the downscaling investigated here is performed repeatedly by the same amount after each task and interleaved with generative replay, which makes the interplay of downscaling in continual learning hard to generalize from previous pruning-based model compression studies.

Our computational experiments identified a tradeoff with the level of downscaling - the maximum early task accuracy is increasingly protected with more downscaling, but more downscaling can degrade performance on more recent tasks. An observable effect of downscaling on model performance started at levels greater than 50\%. At downscaling levels greater than 90\%, performance on recently learned tasks began to be severely degraded. For intermediate downscaling levels of 75\%, the prolonged protection of earlier tasks coincides with the rise of current task accuracy for the highest overall accuracy during training.

This highest overall accuracy is observed early on in the training iterations for an individual task. 
The decrease in average accuracy as training continues during a task can be attributed to the tradeoff in optimizing performance on current vs. replayed tasks. 
Thus, for a deployed system to have the highest continual learning accuracy, training would need to be stopped early for the current task. Note, however, that this higher accuracy cannot be achieved simply by training all tasks for less iterations, because the current set of classes that are learned during a task need to be trained over the full course of iterations to enable future accurate identification.
Interestingly, even though performance drops on a task over training, it increases again (is recovered) (see Fig. \ref{figure.fig9}), which is more pronounced with higher levels of downscaling. This suggests that with higher levels of downscaling, network weights are in a configuration to enable recovery of early task accuracy, even when early task accuracy is diminished.
When investigating weight distributions during this re-learning / higher accuracy period, balanced accuracy starts to diminish as weights become less bi-modal.  
Overall, the interplay between integrating synaptic downscaling with generative replay shows how high (but not extreme) levels of downscaling can be beneficial for continual learning.

While the overall accuracy increased during training with the modeled tripartite artificial sleep, there are ways that the current approach can be extended in conjunction with other approaches for continual learning.
In this work, the synaptic downscaling is used in conjuction with aspects of brain-inspired generative replay \cite{van2020brain}, however there are other generative approaches \cite{hayes2021replay} which could be explored in tandem with synaptic downscaling. 
Perhaps the most similar work to the inclusion of synaptic downscaling for continual learning are pruning-based approaches \cite{golkar2019continual}, which prune neurons based on activation level from earlier tasks in order to compress the current task's model and then iteratively train, progressively utilizing more of the network's capacity with additional tasks. 
Related to pruning-based approaches which protect a subset of  weights completely, are weight regularization approaches, which in effect offers varying levels of "protection" to the weights in a network \cite{kirkpatrick2017overcoming,zenke2017continual}.
With weight regularization approaches, changes to certain weights are protected from subsequent change based on a calculated importance metric, as opposed to the protection of weights based on their magnitude implemented here during synaptic downscaling.
Additionally, with many weight regularization approaches, once a synapse has importance attributed to it for a certain task, that importance will never decrease.
While this is helpful for continual learning to prevent catastrophic forgetting and has been shown to increase accuracy when combined with generative (brain-inspired) replay \cite{van2020brain}, protecting changes to a network at the individual weight level may make the network less likely to reconfigure / re-consolidate larger configurations of weights.
It is interesting that, here, even with pruning 90\% of the weights for each task (and not explicitly calculating an importance), early task performance can be protected quite well.  Our results also suggest testable hypotheses in cognitive neuroscience.  For example, in subjects who learn different memory tasks on each of several days, our model predicts that synaptic downscaling measurements and interventions may correlate with the retention of early learning.  This would of course necessitate tools for the reliable noninvasive measurement of downscaling.



In addition to extensions of existing continual learning approaches, there is an opportunity to guide neuroscience investigations based on observed trends here and to build continual learning models that capture even more aspects of sleep.
In particular, there are several outstanding questions around synaptic downscaling during sleep which could be investigated further.
First of all, a better quantification of size-dependant downscaling during sleep in future experimental studies could help parameterize more detailed models of synaptic downscaling. We performed a magnitude-based zeroing of weights as a first-order approximation of size-dependant downscaling that could be extended in further analysis.
Furthermore, the investigation of how synpatic downscaling changes between neural regions (e.g. hippocampal, cortical sensory, cortical associational) could be integrated into future models as well. We implemented the same amount of downscaling throughout all encoder, decoder, and classifier output layers which could be augmented in future models.

Our implementation of NREM veridical replay could also be augmented in the future.
For simplicity, we replayed intermediate layer representations of the current task's image inputs during model training. This NREM veridical replay mechanism could instead more closely model hippocampal processes, incorporating an additional veridical generator component.

In conclusion, there is a rich panoply of benefits and possible algorithmic extensions suggested by the inclusion of multiple sleep processes (here, three) in the construction of artificial neural networks and intelligent systems. Much of this will benefit from adopting a perspective that includes not only the overt behavioral neuroscience of wakefulness, but also the roughly one-third of our lives spent processing information during sleep.





\bibliographystyle{./IEEEtran}
\bibliography{./IEEEabrv,./references}

\vspace{12pt}

























\end{document}